\documentclass{article}

\usepackage[final, nonatbib]{neurips_2021}

\usepackage[utf8]{inputenc} 
\usepackage[T1]{fontenc}    
\usepackage{hyperref}       
\usepackage{url}            
\usepackage{booktabs}       
\usepackage{amsfonts}       
\usepackage{nicefrac}       
\usepackage{microtype}      
\usepackage[english]{babel}
\usepackage{graphicx}
\usepackage{biblatex}
\usepackage{amsmath}
\usepackage{bbm}
\usepackage{qtree}

\usepackage[table,xcdraw]{xcolor}

\usepackage{graphicx}

\usepackage{fancyhdr}
\usepackage{color}
\usepackage{url}
\usepackage{multirow}
\usepackage{graphicx} 
\usepackage{times}
\usepackage{latexsym}
\usepackage{multicol}
\usepackage{multirow}

\usepackage{tabularx} 
\usepackage{amssymb}
\usepackage{amsmath}
\usepackage{mathtools}
\usepackage{pifont}

\usepackage{ragged2e}

\usepackage{etoolbox}
\makeatletter
\preto{\@tabular}{\parskip=3pt}
\makeatother

\usepackage{enumitem}
\setlist[itemize]{leftmargin=*}

\title{A Closer Look at Advantage-Filtered Behavioral Cloning in High-Noise Datasets}

\hypersetup{
    colorlinks=true,
    linkcolor=blue,
    filecolor=magenta,      
    urlcolor=cyan,
}
\begin{document}

\author{
  Jake Grigsby \\
  University of Virginia \\
  \texttt{jcg6dn@virginia.edu}
  \And Yanjun Qi \\
  University of Virginia \\
  \texttt{yanjun@virginia.edu}
 }

\maketitle

\begin{abstract}
Recent Offline Reinforcement Learning methods have succeeded in learning high-performance policies from fixed datasets of experience. A particularly effective approach learns to first identify and then mimic optimal decision-making strategies. Our work evaluates this method's ability to scale to vast datasets consisting almost entirely of sub-optimal noise. A thorough investigation on a custom benchmark helps identify several key challenges involved in learning from high-noise datasets. We re-purpose prioritized experience sampling to locate expert-level demonstrations among millions of low-performance samples. This modification enables offline agents to learn state-of-the-art policies in benchmark tasks using datasets where expert actions are outnumbered nearly $65:1$. 
\end{abstract}

\section{Introduction}
Reinforcement Learning (RL) promises to automate decision-making and control tasks across a wide range of applications. However, there are many real-world problems for which the typical learning and exploration process is too expensive (e.g., robotics, finance) or too dangerous (e.g., healthcare, self-driving) to make RL a viable solution. Fortunately, there is hope that we can trade RL's reliance on exploration and online feedback for a new reliance on large fixed datasets, much like those that have driven progress in supervised learning over the last decade \cite{russakovsky2015imagenet, levine2020offline, levineOfflineIAS}. In many situations, we can collect experience by recording the decisions of human demonstrators or existing control systems. For example, we can collect data for self-driving systems by mounting cameras and other sensors onto human-driven vehicles. However, as the scale of the dataset collection process grows, it becomes increasingly impractical to verify the quality of the data we are collecting. Even worse: the task may be challenging enough that no quality demonstrations could ever exist; instead, fleeting moments of optimal decision-making occur by chance in a sea of noise collected by a large number of agents. 

Our goal is to build agents that can learn to pick out valuable information in large, unfiltered datasets containing experience from a mixture of sub-optimal policies. Recent work has proposed new ways to mimic the best demonstrations in a dataset \cite{nair2020accelerating, wang2020critic, wang2018exponentially, peng2019advantageweighted, chen2020bail}. 
In this paper, we analyze and address some of the problems that arise in determining which demonstrations are worth learning. We build a custom dataset of experience from popular continuous control tasks to demonstrate the challenge of learning from datasets with few or zero successful trajectories and provide an easy-to-implement solution that lets us learn expert policies from millions of low-performance samples.

\section{Background and Related Work}

In the interest of space, we assume the reader is familiar with Deep RL, continuous control tasks, and the general off-policy actor-critic framework. For an introduction, please refer to \cite{lillicrap2015continuous} and \cite{fujimoto2018addressing}.

\subsection{Offline Reinforcement Learning}
Offline (or ``Batch") RL is a subfield that deals with the special case of learning from static demonstration datasets. In offline RL, our goal is to discover the best policy given a dataset of fixed experience without the ability to explore the environment. The naive application of off-policy RL methods to the fully offline setting often fails due to distribution shift between the dataset and test environment \cite{levine2020offline}. Solutions based on approximate dynamic programming are also vulnerable to exploding Q-values due to the propagation of overestimation error with no opportunity for correction in the online environment \cite{kumar2019stabilizing}.

Modern approaches to offline RL broadly involve modifications to standard off-policy algorithms to reduce overestimation and minimize distributional shift. REM uses an ensemble of critics to reduce overestimation error \cite{agarwal2020optimistic}. UWAC \cite{uwac} uses the uncertainty of value predictions to reduce error propagation. CQL \cite{kumar2020conservative} penalizes out-of-distribution Q-values to encourage in-distribution actions. MOPO \cite{yu2020mopo} is a model-based method that penalizes actions that leave the data distribution that the model was trained on.

\subsection{Advantage-Filtered Behavioral Cloning}
While much progress has been made in adapting off-policy RL to the constraints of the offline setting, these methods often struggle to outperform simple Behavioral Cloning (BC) \cite{qin2021neorl, fu2021d4rl}. In BC, we train an actor network to replicate the actions taken in the demonstration dataset:

\begin{align}
    \mathcal{L}_{actor} = \mathop{E}_{(s, a) \sim \mathcal{D}}\left[-\text{log} \pi_{\theta}(a \mid s)\right]
\label{bc_loss}
\end{align}

Behavioral Cloning's inability to outperform its demonstrations becomes a major shortcoming when the dataset contains sub-optimal experience. Ideally, we would discard trajectories that contain low-quality demonstrations and only clone the best behavior available. This is the core idea behind Filtered Behavioral Cloning (FBC). FBC compares demonstrations to discard or down-weight the advice of sub-optimal policies. A reasonable metric for comparison is the advantage function, defined as $A^{\pi}(s, a) = Q^{\pi}(s, a) - V^{\pi}(s)$, which represents the change in expected return when taking action $a$ instead of following the current policy. Filtering experience by advantage is essentially a counterfactual query --- we compute how much better action $a$ \textit{would have been} than the action chosen by our policy. We will refer to BC methods that filter experience by advantage as Advantage-Filtered Behavioral Cloning (AFBC). Because we do not have access to the true $Q^{\pi}$ or $V^{\pi}$ functions, we need to estimate the advantage. There are two main approaches:
\begin{enumerate}
    \item \textbf{Monte Carlo Advantages} define an estimate of the advantage $\hat{A}^{\pi}(s, a) = \hat{\eta}(s, a) - V_{\phi}(s)$, where $\hat{\eta}(s, a)$ represents the empirical expected return of trajectories in the buffer that contain this $(s, a)$ pair and $V_{\phi}(s)$ represents a learned state value function parameterized by a neural network with weights $\phi$.
    \item \textbf{Q-Based Advantages} define an estimate of the advantage $\hat{A}^{\pi}(s, a) = Q_{\phi}(s, a) - \mathop{E}_{a' \sim \pi(s)}[Q_{\phi}(s, a')] \approx Q_{\phi}(s, a) - \frac{1}{k}\mathop{\Sigma}_{i=0}^{k}Q_{\phi}(s, a' \sim \pi_{\theta}(s))$. We typically learn $Q_{\phi}$ as in standard off-policy actor-critics. The BC loss prevents the actor policy from diverging from the demonstration policy and reduces the risk of overestimation error. 
\end{enumerate}

Once we have a reliable estimate of the advantage function, we add a ``filter" to the BC loss (Eq \ref{bc_loss}):

\begin{align}
\label{afbc_loss}
    \mathcal{L}_{actor} = \mathop{E}_{(s, a) \sim \mathcal{D}}\left[-f(\hat{A}^{\pi}(s, a))\text{log} \pi_{\theta}(a \mid s)\right]
\end{align}

There are two common choices of $f$:
\begin{enumerate}
    \item \textbf{Binary Filters} define $f(\hat{A}(s, a)) = \mathbbm{1}_{\{\hat{A}^{\pi}(s, a) > 0\}}$. This creates a boolean mask that eliminates samples which are thought to be worse than the current policy \cite{wang2020critic}.
    \item \textbf{Exponential Filters} define $f(\hat{A}(s, a)) = \text{exp}(\beta \hat{A}^{\pi}(s, a))$, where $\beta$ is a hyperparameter. The advantage values are often clipped or normalized for numerical stability \cite{wang2018exponentially, nair2020accelerating}.
\end{enumerate}

  The methods implemented and expanded upon in this paper are most directly related to CRR \cite{wang2020critic} and AWAC \cite{nair2020accelerating}, as discussed in Section \ref{implementation}. AWAC and CRR are roughly concurrent publications that arrive at a very similar method. CRR is focused on offline RL in the RL Unplugged benchmark \cite{fu2021d4rl}, while AWAC is more concerned with accelerating online fine-tuning by pre-training on smaller offline datasets. These methods are the latest iteration of the core AFBC idea that has appeared many times in recent literature. This section is partly an attempt to unify the literature and highlight a common theme that has been somewhat under-recognized. To the best of our knowledge, the first deep-learning-era AFBC implementation is MARWIL \cite{marwil}, which performed BC on samples re-weighted by their Monte Carlo advantage estimates and applied the technique to environments like robot soccer and multiplayer online battle arenas. AWR \cite{peng2019advantageweighted} is a very similar method that focuses on continuous control tasks. BAIL \cite{chen2020bail} learns an upper-envelope value function that encourages optimistic value estimates, and therefore conservative advantage estimates. Samples are filtered based on heuristics of their advantage relative to the rest of the dataset (e.g. with advantage larger than $x$ or larger than $x\%$ of all samples, where $x$ becomes a hyperparameter). A similar approach appears in the experiments of Decision Transformer \cite{chen2021decision}. Nair et al. \cite{nair2018overcoming} accelerate off-policy learning by providing an expert action when its value exceeds that of the agent's policy (according to the critic network) - essentially creating a deterministic binary filter. SIL \cite{oh2018selfimitation} combines an AFBC-style loss with the standard policy gradient in an online setting, where the BC step greatly improves sample efficiency. SAIL \cite{ferret2020self} adds Q-based advantages to SIL and corrects outdated MC estimates by replacing them with parameterized estimates when they become too pessimistic. Gangwani et al. \cite{gangwani2019learning} give another approach to self-imitating policy gradients for sample-efficient (online) continuous control. SQIL \cite{reddy2019sqil} and ORIL \cite{zolna2020offline} use standard off-policy optimization with a modified reward function that encourages the agent to stay in (or return to) $(s, a)$ pairs that are covered by the dataset, thereby helping it recover from distributional shift. We provide a rough classification of the most relevant modern literature according to the tree diagram in Figure \ref{afbc_tree}.

Imitation Learning \cite{Osa_2018} also deals with learning from sub-optimal demonstrations. While the goals and experiments are similar to the offline RL setting, the methods are typically quite different. See VILD \cite{pmlr-v119-tangkaratt20a}, IC-GAIL, and 2IWIL \cite{wu2019imitation} for recent examples.

\begin{figure}[t]
\centering
\Tree [.Advantage-Filtered\ Behavioral\ Cloning [.Monte\ Carlo [.Binary \textbf{BAIL\ \cite{chen2020bail}} ][.Exponential \textbf{AWR\ \cite{peng2019advantageweighted}} \textbf{SIL$^\star$\ \cite{oh2018selfimitation}} \textbf{MARWIL\ \cite{marwil}} ]][.Q-Based [.Binary \textbf{CRR\ Binary\ \cite{wang2020critic}} \textbf{OERL\ \cite{nair2018overcoming}} ][.Exponential \textbf{CRR\ Exp\ \cite{wang2020critic}} \textbf{SAIL$^\star$\ \cite{ferret2020self}} \textbf{AWAC\ \cite{nair2020accelerating}} ]]]
\caption{A rough classification of the recent Advantage-Filtered Behavioral Cloning literature, which contains a surprising amount of overlap due to concurrent publication and changes in motivation/experimental focus. $^\star$SIL and SAIL use a clipped rather than an exponential advantage, but the loss is still decreasing in the magnitude of the advantage.}
\label{afbc_tree}
\end{figure}

\section{Datasets for High-Noise Offline RL}
Our goal is to investigate the performance of AFBC as the quality of the dataset decreases, and useful demonstrations are hidden in sub-optimal noise. The first step is to gather those fixed datasets and create a systematic approach to varying the quantity and quality of demonstrations provided to our agents. Learning from sub-optimal data is a widely recognized challenge for offline RL, and as such it is a key component of recent benchmarks, including D4RL \cite{fu2021d4rl}, RL Unplugged \cite{gulcehre2021rl}, and NeoRL \cite{qin2021neorl}. However, these benchmarks do not give us enough control over the demonstration dataset for our purposes. Of these, NeoRL and D4RL are closest to what we are looking for - both offer low, medium, and expert performance datasets of varying sizes but are not large enough for some of our experiments and only offer those three tiers of quality. In contrast, RL Unplugged contains enormous datasets dumped from agent replay buffers throughout training but gives us less control over the quality of any given subset of that data. In an effort to get the best of both worlds, we collect our own offline RL dataset in the following manner:

\begin{figure}[t]
    \centering
    \includegraphics[width=\linewidth]{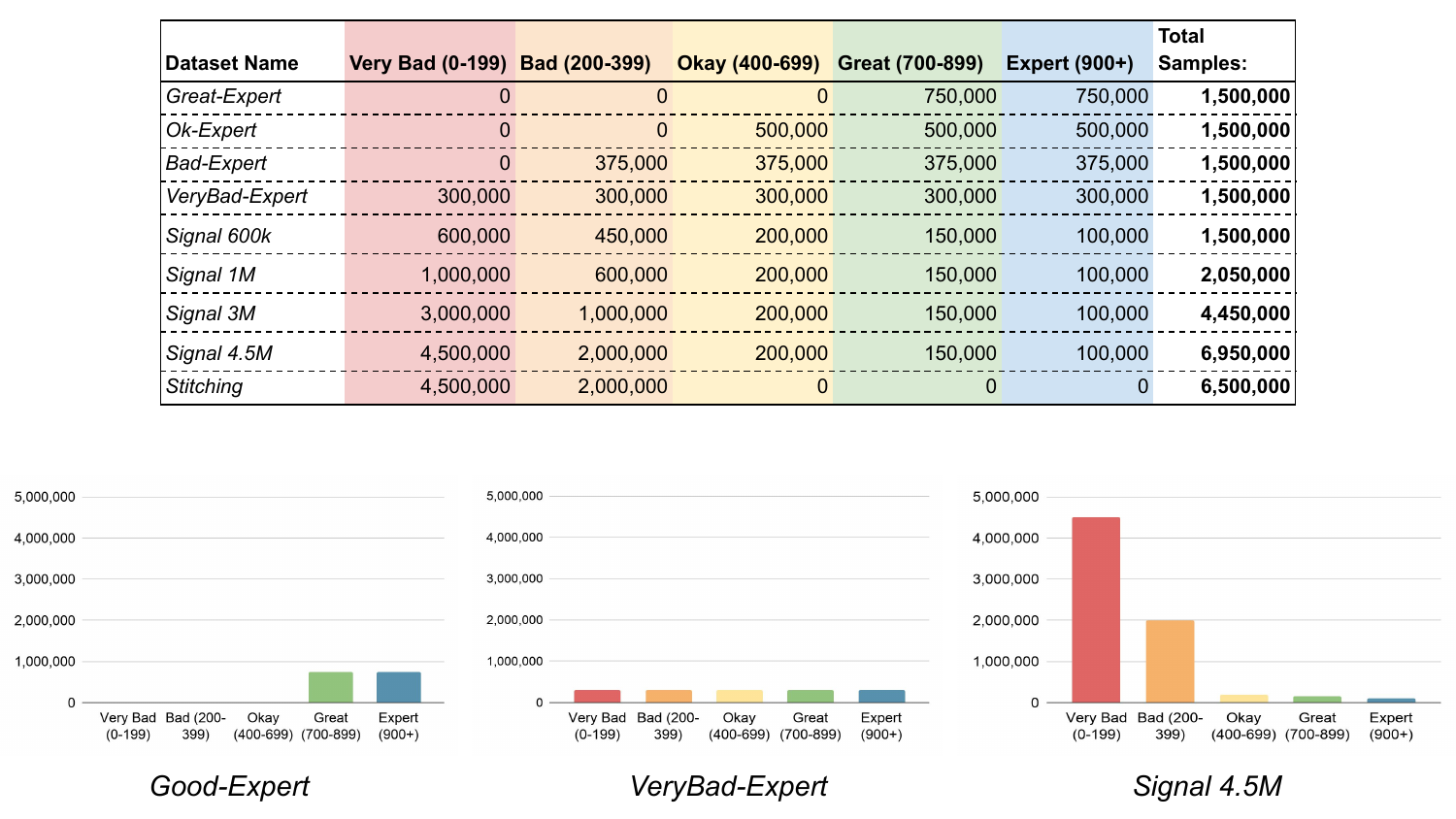}
    \caption{\textbf{Distribution of samples in all $9$ dataset types.} We begin with near-optimal BC data and slowly shift the distribution towards low-performance noise.}
    \label{fig:tasks}
\end{figure}

\begin{enumerate}
    \item We train online agents in $5$ tasks from the DeepMind Control Suite. The D4RL and RL Unplugged authors stress the importance of diverse policies and state-space coverage. With that in mind, we train multiple policies per environment using a mixture of algorithms and hyperparameters. More specifically, we use (deterministic) TD3 \cite{fujimoto2018addressing} and several variants of (stochastic) SAC \cite{haarnoja2018soft}, implemented in \cite{deepcontrol}.
    \item We pause at regular intervals during training and record the policies' behavior for $50$ episodes of environment interaction. To broaden our state-action space coverage, we sample from stochastic policies (rather than taking the mean, as is standard for test-time policy evaluation). The average undiscounted return of the policy is saved alongside the ($s$, $a$, $r$, $s'$, $d$) tuples.
    \item The saved experience is split into $5$ levels of performance. The DMC tasks have returns in the range $0 - 1000$, and we divide the dataset according to the average return of the policy at the time they were recorded. The final result is a large dataset with millions of samples and five levels of performance. Full size and quality information is listed in Appendix \ref{dset-details} Figure \ref{tbl:dset_splits}.
\end{enumerate}

From this raw data, we create a series of $9$ offline RL datasets for each task. The size and makeup of each dataset is listed in Figure \ref{fig:tasks}. \textit{Great-Expert} splits $1,500,000$ samples evenly across expert and high-performance demonstrations. This represents a near best-case scenario for behavioral cloning methods in which very little data needs to be actively ignored. \textit{Okay-Expert} begins a trend where the same $1,500,000$ sample budget is allocated to lower quality data, making vanilla BC less and less effective. This concludes with \textit{VeryBad-Expert}, where samples are split evenly across all $5$ performance bins. The next $4$ datasets limit expert experience to just $100,000$ samples, which our experiments show is typically enough to train successful policies without overfitting. The challenge is finding those $100,000$ samples in a buffer of millions of poor demonstrations. The most difficult of these is \textit{Signal 4.5M}, where sub-optimal data crowds the dataset at a ratio of almost $65:1$. Finally, we test the agent's ability to extract knowledge from pure noise. In the \textit{Stitching} task, the agent is given $6,500,000$ samples of experience from agents that perform only slightly better than random policies and must learn to identify the few moments of optimal decision making and combine them into a single policy. 

We benchmark the challenge of Behavioral Cloning in continuous control tasks by training BC agents on our high-noise and multi-policy datasets. The results are shown in Appendix \ref{additional-results} Figure \ref{fig:sbc_results}. BC often fails to make progress, even with the \textit{Great-Expert} datasets, presumably because of distribution shift or mixed-policy learning. We see a sharp decline in BC performance as the dataset fills with noisy demonstrations.



\section{Challenges in AFBC}
\subsection{AFBC Baseline and Implementation Details}
\label{implementation}
We implement a custom AFBC baseline combining elements of AWAC and CRR. Monte Carlo advantage estimates can work well in practice but limit our ability to learn from low-performance data by creating pessimistic advantage estimates. Therefore, we adopt the Q-Based estimates from CRR and AWAC, using critic networks trained similarly to SAC \cite{haarnoja2018soft} and TD3 \cite{fujimoto2018addressing}. 
\begin{align}
    \mathcal{L}_{critic} &= \mathop{\mathbb{E}}_{(s, a, r, s') \sim \mathcal{D}} \left[\frac{1}{2}\sum_{i=1}^{2}\bigg(Q_{\phi, i}(s, a) - \big(r + \gamma(\mathop{\text{min}}_{j=1,2}Q_{\phi', j}(s', \Tilde{a}'))\big)\bigg)^2\right], \Tilde{a}' \sim \pi_{\theta}(s')
\end{align}

The advantage estimate is computed using $4$ action samples:
\begin{align}
    \hat{A}^{\pi}(s, a) = \frac{1}{2}\sum_{j=1,2}Q_{\phi, j}(s, a) - \frac{1}{4}\sum_{i=0}^{4}\bigg(\frac{1}{2}\sum_{j=1,2}Q_{\phi, j}(s, a' \sim \pi_{\theta}(s))\bigg)
\end{align}
Note that we do not use distributional critics as in CRR - nor do we use the the critic weighted policy technique at test time. The policy is a tanh-squashed Gaussian distribution \cite{pytorch_sac}.  We use a single codebase for all experiments in order to control for small implementation details \cite{henderson2019deep}, and have extensively benchmarked our code against existing results in the literature \cite{fu2021d4rl, qin2021neorl}. See Appendix \ref{appendix-implementation} 
for more implementation details and a full list of hyperparameters.

\subsection{Binary vs. Exponential Filters}
While the exponential filter has been used with great success in prior work, it comes with two non-trivial implementation challenges. First, we need to deal with the magnitude of advantages across different environments. Advantage estimates are dependent on the scale of rewards, which can vary widely even across similar tasks in the same benchmark (e.g., Gym MuJoCo \cite{brockman2016openai}). There are plenty of reasonable approaches to solving this. The simplest is to clip advantages in a numerically stable range, but this runs the risk of losing the ability to differentiate between high-advantage actions. MARWIL keeps a running average and normalizes the advantages inside the filter. The AWAC codebase considers several alternatives, including softmax normalization. An interesting alternative is PopArt \cite{vanhasselt2016learning, hessel2018multitask}; by standardizing the output of our critic networks we rescale advantages and get the benefits of PopArt's stability and hyperparameter insensitivity for free. The second challenge is the temperature hyperparameter $\beta$. Prior work demonstrates significant changes in performance across similar $\beta$ values \cite{marwil}, and is often forced to use different settings in each domain \cite{nair2020accelerating}. 

\begin{figure}[t]
    \centering
    \includegraphics[width=\linewidth]{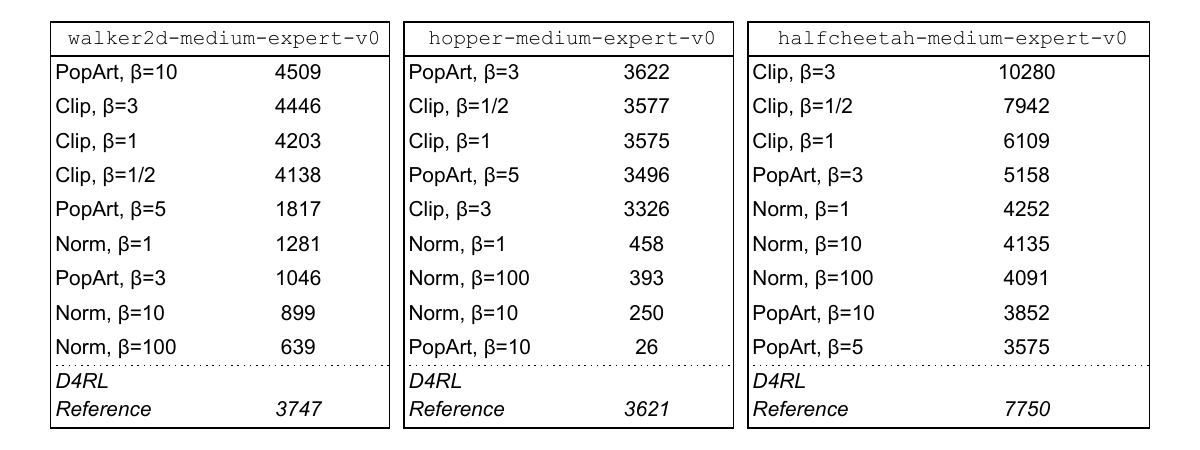}
    \caption{\textbf{The implementation challenges of exponential filters.} We test $9$ exponential-filter variants on three tasks from the D4RL benchmark. The scores are listed in decreasing order; there is little correlation between a filter's rank in one task and its rank in the others. The `D4RL Reference` row lists the highest reported score of any algorithm from the D4RL results \cite{fu2021d4rl}, to provide some context.}
    \label{fig:exp-results}
\end{figure}

We demonstrate the extent of the problem by implementing $9$ reasonable variants and evaluating them on tasks from the D4RL benchmark. We use D4RL instead of our custom datasets in order to validate our implementation and compare against previously published results. The results are shown in Figure \ref{fig:exp-results}. There is very little correlation between a setting's relative return in one task and its performance in the others, making it difficult to set a high-performance default a priori\footnote{Note that all filters in our codebase are clipped for numerical stability such that exponential filters with large $\beta$ values begin to approximate binary filters.}. A change in $\beta$ is enough to take us from near-failure to state-of-the-art performance. We argue that this could be considered a major shortcoming when benchmarking research and dealing with real-world problems, especially in a field that already suffers from significant implementation issues \cite{engstrom2020implementation, henderson2019deep}. For this reason, all following experiments use a binary filter.

\subsection{Advantage Distributions and Effective Batch Size}
\label{problems_list}
In an effort to get a better understanding of the dynamics of the AFBC algorithm, we track the distribution of advantages in the dataset throughout training. Example results on the \textit{VeryBad-Expert} datasets  are shown in Figure \ref{adv_histograms}. Using data from our experiments, we identify three challenges with current AFBC methods:

\begin{figure}[h!]
    \centering
    \makebox[\textwidth][c]{\includegraphics[width=1.2\textwidth]{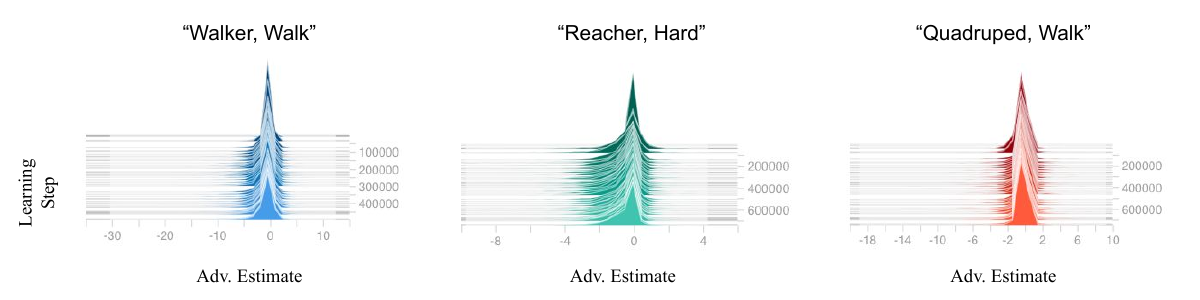}}
    \caption{\textbf{Distribution of advantage (adv.) estimates throughout training.} We estimate the advantage of a large sample of demonstrations from the replay buffer on the \textit{VeryBad-Expert} datasets for three control tasks where AFBC performs well. These figures were generated for every task and dataset combination with similar results.}
    \label{adv_histograms}
    \vspace{1mm}
\end{figure}

\paragraph{Effective Batch Size.} Regardless of dataset quality, there are many samples with negative advantage. In a typical task, the critic networks assign a negative advantage to more than half of the dataset - and that number can rise as high as $90\%$ in some cases. This means that we rely on a small fraction of each batch to compute our gradients and improve the actor. The effect is most apparent when using binary filters, where we may be entirely disregarding a large portion of each minibatch. Exponential filters attribute at least some learning signal to each sample, but we are likely to be down-weighting most of our batch. The AFBC baselines show that learning is still possible at low signal-to-noise ratios, but we are sacrificing stability by reducing our network's effective batch size. In noisy datasets (\textit{Signal-4.5M}), advantageous samples become so rare that a uniform sample of data yields batches with prohibitively high variance.

\paragraph{Noisy Labels.} Advantage estimates are highly concentrated at low absolute values. Many of the demonstrations have advantage estimates that oscillate close enough to zero that they can be labeled as positive or negative based on randomness in the estimator.

\paragraph{Static Advantage Distributions.} The advantage distribution does not change very much over time. The estimator spends the first few thousand learning steps adapting to the task's reward scale but makes few adjustments over the remaining steps. A priori, we might expect the agent to clone most or all of the dataset before becoming more confident in its ability to identify the best strategies. Instead, the estimator quickly learns to clone a small percentage of the dataset.

\subsection{Addressing Effective Batch Size with Prioritized Sampling}
\begin{figure}[th!]
    \centering
    \includegraphics[width=.45\textwidth]{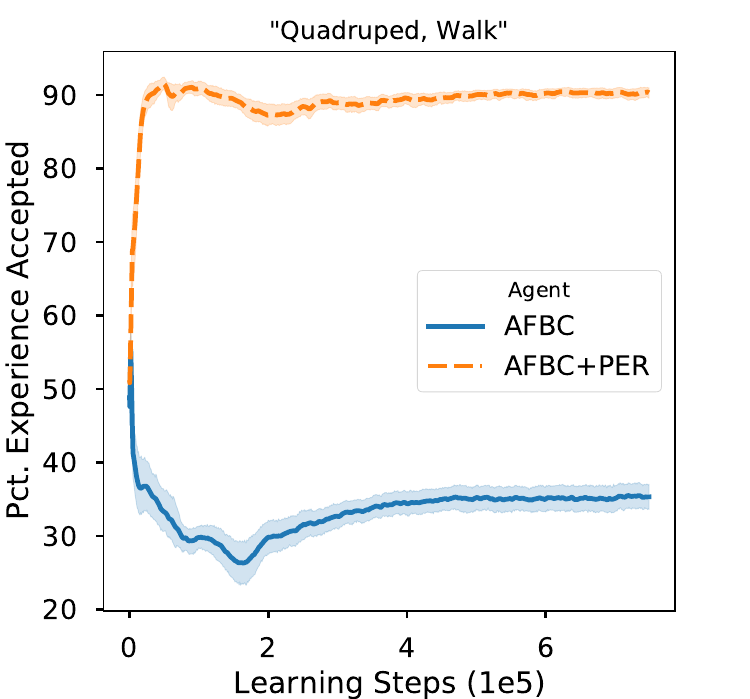}
    \includegraphics[width=.45\textwidth]{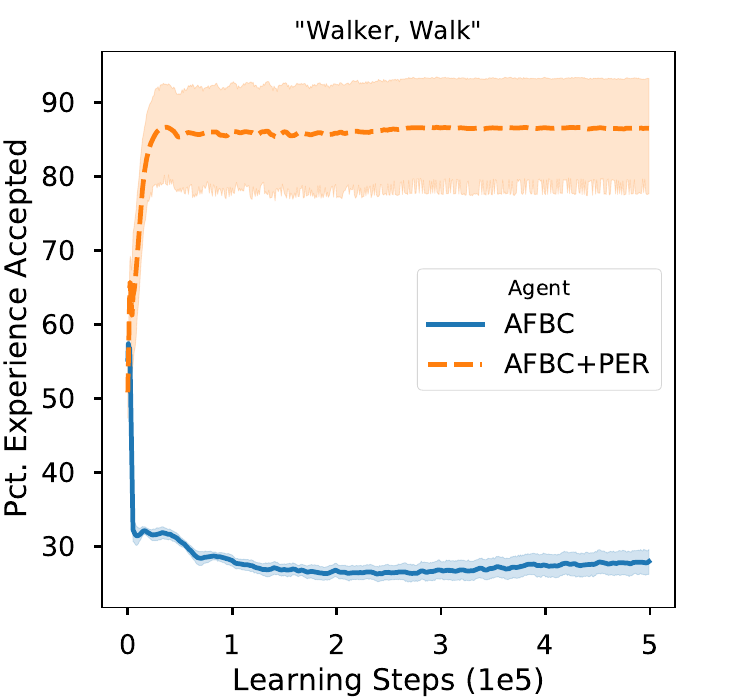}
    \caption{\textbf{Effective batch size of AFBC agents}. We measure the percentage of actor update batches that are approved by a binary filter throughout training. Curves represent the mean and $95\%$ confidence interval across $5$ random seeds in all $8$ datasets that contain expert experience. PER samples actions that are more likely to have positive advantage, meaning that fewer samples are masked or down-weighted by our advantage filter.} 
    \label{fig:experience_curves}
\end{figure}

As discussed above, learning from sub-optimal datasets with low-advantage actions reduces our actor network's batch size. We can improve by sampling batches that are more likely to be accepted by the advantage filter. A straightforward way to implement this is to re-purpose Prioritized Experience Relay (PER) \cite{schaul2016prioritized} to sample high-advantage actions rather than high-error bellman backups. The critic update proceeds as usual - sampling uniformly from the replay buffer and minimizing temporal difference error across the entire dataset. This gives us the opportunity to re-compute advantages and identify new useful demonstrations. During the actor update, we sample transitions from the buffer proportional to the advantage we computed when they were last used to train the critic. We can still filter low-performance samples, but this is much less likely to be necessary (see Figure \ref{fig:experience_curves}). PER also seems to offer a partial solution to the noisy labels problem; for a sample to be presented to the filter inside of the actor update, it must have been assigned a positive advantage at some point in the recent past. Before we clone the action, it then has to be labeled positive \textit{again} - reducing our ability to be fooled by the large group of near-zero-advantage actions.

\paragraph{Prioritized Experience Replay Details: } The Prioritized Experience Replay implementation is based on OpenAI Baselines \cite{baselines}. The replay is given the option of sampling uniformly from the underlying buffer or using prioritized sampling. We sample uniformly during the critic update and then update the samples' priorities according to $\text{max}(\hat{A}^{\pi}(s, a), \epsilon)$, where $\epsilon$ is a small positive constant. We also experiment with binary priority weights ($\mathbbm{1}_{\{\hat{A}^{\pi}(s, a) \geq 0\}} + \epsilon$). See Appendix \ref{additional-results} Figure \ref{fig:weight_comp} for a brief comparison. We do not use any importance sampling weights. This is a key difference between our use-case and the traditional use of PER. Normally, we prioritize samples based on a value that is highly correlated with their effect on the gradient (e.g., absolute TD error) and need to use importance weights to compensate for sampling the highest priorities. This priority system decreases our chances of sampling actions that our filter will discard but does not suffer from the same skewed gradient values. As the actor takes gradient steps in the direction of the approved experience, its action probabilities center around the provided action until the advantage drops to zero. In order to keep track of this effect, we re-compute the priorities of the sampled actions after each actor update.

\begin{figure}
    \centering
    \makebox[\textwidth][c]{\hspace{-10mm}\includegraphics[width=1.3\linewidth]{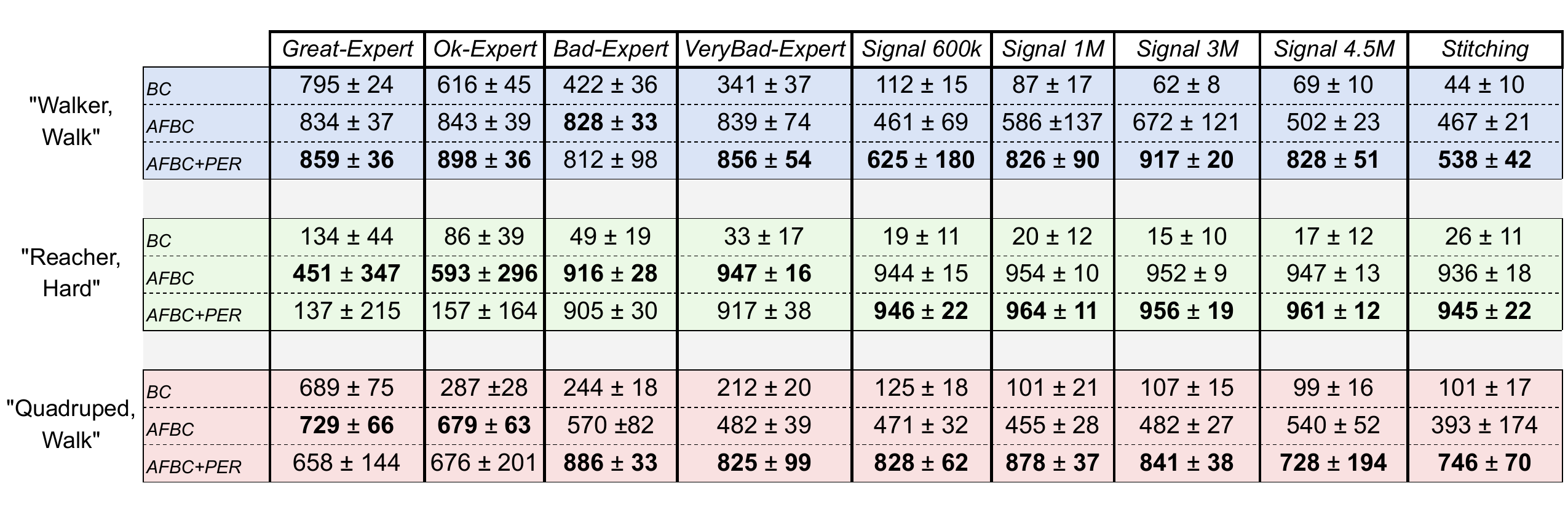}}
    \caption{\textbf{AFBC in High-Noise Continuous Control Tasks.} Prioritized sampling allows AFBC to continue to learn high-performance policies when uniform sampling is overwhelmed by noise. We use a consistent set of hyperparameters for each task, and report the mean and $95\%$ confidence interval across $5$ random seeds. Results on more difficult tasks in Figure \ref{fig:hard_afbc}.}
    \label{afbc_results}
\end{figure}

This simple trick leads to a dramatic improvement in performance on the \textit{Signal} datasets. Results are listed in Figures \ref{afbc_results} and \ref{fig:hard_afbc}. Simply put: if AFBC can learn a high-quality policy, AFBC with the PER trick can maintain that performance despite an enormous amount of noise. This technique also increases performance in the \textit{Stitching} task, thanks to its enhanced ability to identify the rare optimal decisions of random or near-random policies. It also risks overfitting and instability on the high-quality datasets; this can be corrected by adjusting the $\alpha$ prioritization parameter of the experience replay. Low $\alpha$ values reduce PER to uniform sampling. These experiments use $\alpha = .6$.

\subsection{Learning from Worst-Case Experience}
\label{mc1d-sec}
\begin{figure}[h!]
    \centering
    \makebox[\linewidth][c]{\includegraphics[width=1.2\linewidth]{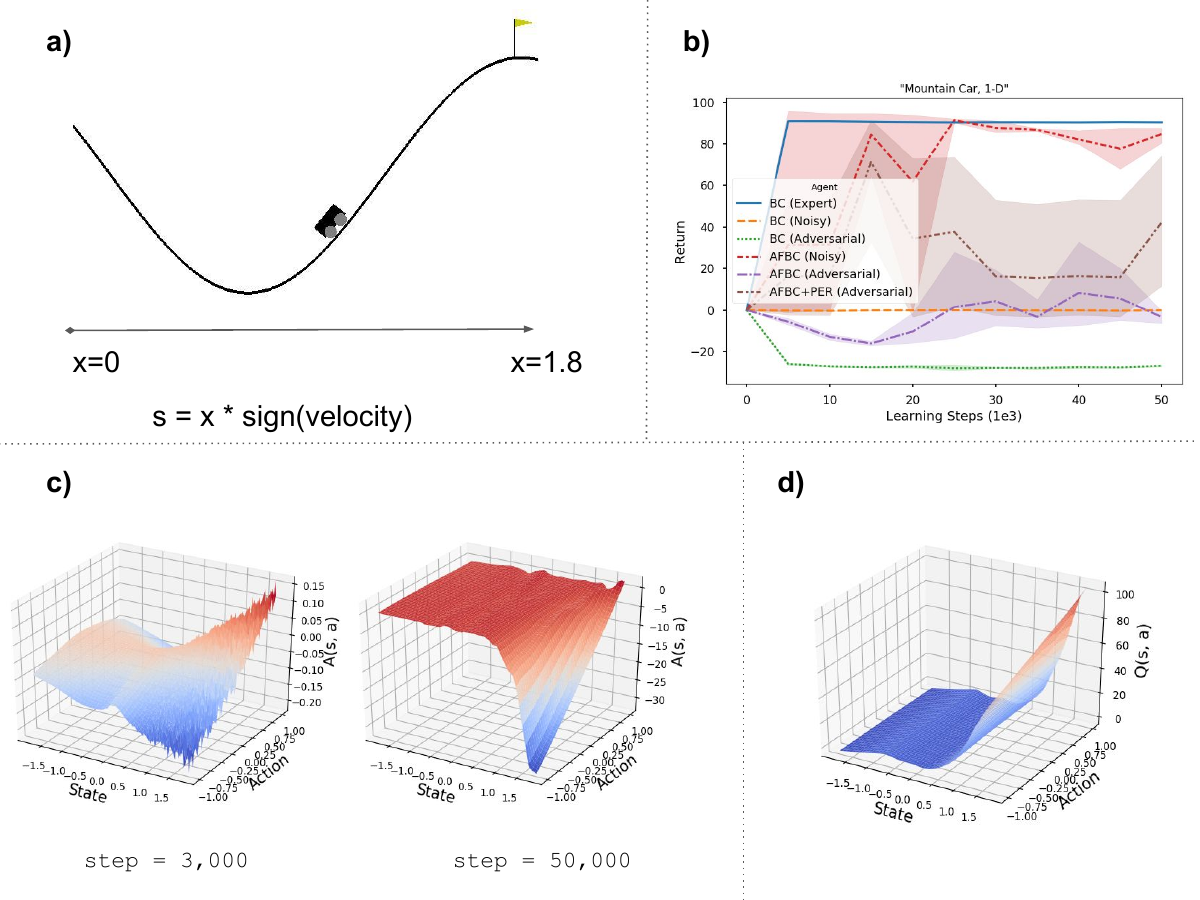}}
    \caption{\textbf{Adversarial Demonstrations in Modified Mountain Car.} a) We create a version of the Mountain Car task that combines state and velocity information into a scalar state that is convenient for plotting. b) AFBC can solve the task despite a $9:1$ ratio of random actions and can even learn despite a dataset of worst-case demonstrations. c) We show the evolution of the learned advantage function across the full $(s, a)$ space. d) The final Q-function.}
    \label{fig:mc1d}
    \vspace{2mm}
\end{figure}

One question that arises when working with large sub-optimal datasets is what may happen in the worst-case scenario where much of the dataset is intentionally misleading. In this setting, we lose the ability to stitch together random policies and must instead learn to isolate the expert data and ignore everything else. To research this situation, we need a domain where humans can easily intuit about the optimal strategy and identify the worst possible policy. Therefore, we will take a brief detour from the high-dimensional robotic control tasks of the DMC Suite and consider the classic ``Mountain Car" task in which an under-powered car learns to gain momentum by going backward to summit a large hill. The environment is pictured in Figure \ref{fig:mc1d}a. The agent receives a large positive reward $+100$ for reaching the goal flag on top of the mountain, with a small penalty for fuel expenditure along the way. The worst possible solution is to gain the speed necessary to climb up the mountain before deciding to turn around and return to the starting position, thereby wasting as much fuel as possible. The default state space is a $(position, velocity)$ tuple, but we compress this information to a scalar representing the current position and direction of movement to plot the $Q$ and $A$ functions as a 3D surface. The action space is bounded in $[-1, 1]$, where positive actions accelerate the cart to the right, and negative actions accelerate it to the left. 

The compressed state space makes the problem more difficult by hiding the velocity information necessary to manage fuel expenditure, but it is still possible to reach the goal flag. We collect a dataset of expert TD3 actions, alongside actions from a random policy. We also isolate random actions that display worst-case behavior where the cart is making progress up the mountain but reverses to move downhill. We create three offline datasets: expert demonstrations, a $9:1$ ratio of random and expert demonstrations, and a $9:1$ ratio of worst-case (or `adversarial') actions. Figure \ref{fig:mc1d}b shows the results of BC, AFBC and AFBC+PER on these datasets. BC performs as expected: it learns from expert data but is distracted by random actions and is confused by the adversarial demonstrations. Default AFBC can ignore the random actions but still succumbs to the adversarial advice.  We attribute this to Q-value inflation - a hypothesis discussed in the appendix of the CRR \cite{wang2020critic} paper; in short: the adversarial advice is so concentrated in a small region of $(s, a)$ space that the bias of consistent Q-updates causes the critic to overestimate the advantage of the adversarial actions. AFBC with the PER trick is better equipped to handle this situation. By replaying experience proportional to its advantage, PER ignores actions that have a small positive advantage due to bias and clones more of the true expert data. AFBC+PER can solve the task based on the adversarial dataset, although the sparse reward has inherently high variance. The critic networks correctly minimize the advantage of the adversarial data (Figure \ref{fig:mc1d}c) and learn an accurate Q-function (Figure \ref{fig:mc1d}d).

\section{Discussions \& Conclusion}
\vspace{-5pt}
The ability to automatically learn high-performance decision-making systems from large datasets will open up exciting opportunities to safely and effectively apply Reinforcement Learning to the real world. There are many domains where we can find enough data to train large neural network policies but cannot verify the demonstrated actions' quality. Success hinges on our ability to answer counterfactual questions about the data: are the decisions made in the dataset the correct ones, or can we find a way to do better? Advantage-Filtered Behavioral Cloning offers an intuitive way to formalize and answer this question. In this paper, we have conducted a thorough empirical study that attempts to unify existing techniques, identify critical obstacles, and provide assurance that this method can learn from unfiltered datasets of any size. However, the prioritized sampling method does not fully address the noisy label and static distribution problems discussed in Section \ref{problems_list}. We experimented with several theoretical solutions that provided somewhat underwhelming improvements on our baseline tasks. Please see Appendix \ref{future_dirs} for a thorough discussion of future directions. 

Methods that can ignore or even improve using low-quality data are valuable because they simplify offline RL dataset collection by reducing the risk that additional data will damage the system. Adding more data is rarely unhelpful and is likely to increase performance. This creates an engineering situation similar to Deep Supervised Learning, where more data and a bigger model are never the wrong answer. Moving forward, we hope to combine this approach with the kinds of large network architectures and high-dimensional datasets that have spurred progress in Deep Supervised Learning to solve complicated tasks beyond simulated control benchmarks.


\printbibliography

\appendix
\section{Dataset Details}
\label{dset-details}
\begin{figure}[h!]
    \centering
    \includegraphics[width=.7\linewidth, height=150px]{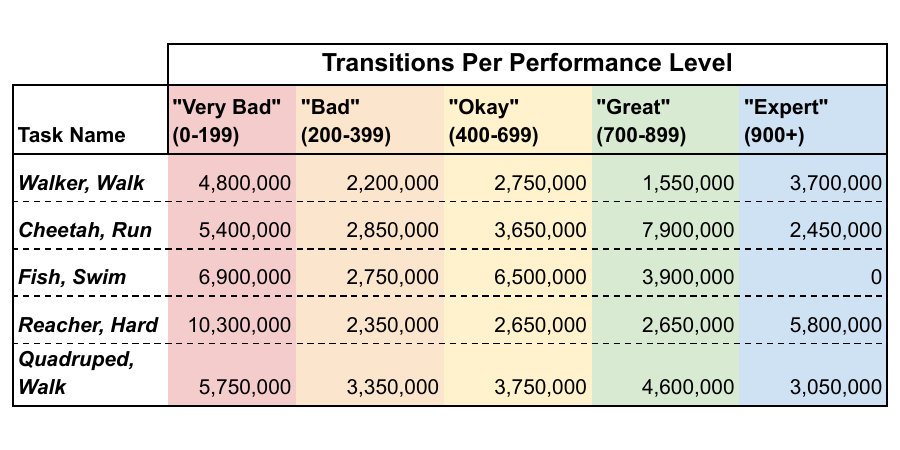}
    \caption{\textbf{Available dataset sizes.} Quantity of Q-learning samples available for each performance range across the $5$ control tasks studied.}
    \label{tbl:dset_splits}
\end{figure}

We provide a listing of the quantity of available Q-learning samples for each environment in Figure \ref{tbl:dset_splits}. Figure \ref{fig:tasks} shows the breakdown of how samples are distributed to generate datasets of varying quality.

\section{Implementation Details}
\label{appendix-implementation}

The code for the experiments in this paper can be found at \textcolor{blue}{\url{https://github.com/jakegrigsby/cc-afbc}}. An updated implementation of our preferred binary-filter PER variant (along with standard online learning and a number of other training tricks from the literature) can be found at \textcolor{blue}{\url{https://github.com/jakegrigsby/super_sac}}.

\subsection{Advantage Weighted Actor-Critic and Critic Regularized Regression Baselines}

Hyperparameters and implementation details are listed in Table \ref{hparams_baseline}.

\begin{table}[h]
\centering
\begin{tabular}{|l|l|}
\hline
\textbf{Policy Log Std Range} & {[}-10, 2{]}           \\
\textbf{Target Delay}         & 2                      \\
\textbf{Weight Decay}         & None                   \\
\textbf{Gradient Clipping}    & None                   \\
\textbf{Actor LR}             & 1e-4                   \\
\textbf{Critic LR}            & 1e-4                   \\
\textbf{Eval Interval}        & 5000                   \\
\textbf{Eval Episodes}        & 10                     \\
\textbf{Buffer Size}          & Size of Dataset        \\
\textbf{Gamma}                & 0.99                   \\
\textbf{Tau}                  & 0.005                  \\
\textbf{Batch Size}           & 512                    \\
\textbf{Gradient Steps}       & 750k                   \\
\textbf{Max Episode Steps}    & 1000                   \\
\textbf{Action Bound}         & {[}-1, 1{]}            \\
\textbf{Network Architecture}         & (256, ReLU, 256, ReLU) \\
\hline
\end{tabular}
\caption{Default hyperparameters for the main DMC experiments.}
\label{hparams_baseline}
\end{table}

We use a tanh-squashed Gaussian distribution for the actor, implemented as in \cite{pytorch_sac}. We also ran trials on every dataset with the implementation in \cite{SpinningUp2018}, a custom implementation of the Beta distribution described in \cite{pmlr-v70-chou17a}, as well as some shorter runs with several other publicly available implementations. The results were surprisingly mixed. The distribution parameterization can be a critical implementation detail because unlike standard online approaches, the AFBC algorithm often requires us to compute log probabilities of foreign actions far outside the center of our actor's own distribution. Numerically stable log prob computations are challenging, and implementations designed for online algorithms may have had no reason to test for stability in this context. Even successful implementations typically return large log prob values early in training. This is not an issue as long as they quickly stabilize to a reasonable range, and we use some protective clipping (e.g. at clearly unstable values like $\text{log}\pi_{\theta}(a) \in (-1000, 1000)$) to minimize the damage.

\subsection{PopArt}

PopArt is implemented as described in \cite{vanhasselt2016learning} and \cite{hessel2018multitask}. We use an adaptive step size when computing the normalization statistics in order to reduce reliance on initialization. The $\nu$ value is initialized to a large positive constant to improve stability (see \cite{vanhasselt2016learning} Pg 13).

\subsection{Evaluation}
The scores displayed in the tables above are computed by:
\begin{enumerate}
    \item Smoothing the learning curve with a polyak coefficient of $.65$.
    \item Determining the mean and standard deviation of several (usually $5$) smoothed learning curves from different random seeds. This creates a lower-variance learning curve.
    \item Reporting the average return and two standard deviations of our low-variance learning curve over the last $10$ evaluations of training.
\end{enumerate}

\section{Alternative Binary Filters, Enhanced Critic Updates, and Future Directions}
\label{future_dirs}
While prioritized sampling is an effective solution to small effective batch sizes, we would still like to learn more accurate advantage estimates and dynamic acceptance curves. In this section, we describe several enhancements in hopes of furthering future research.

\paragraph{Annealed Binary Filters with Statistical T-Tests.} Binary filters appear to rush towards an experience approval percentage that is overly pessimistic early in training. One way to address this is to introduce an overly optimistic filter and adjust its approval criteria during the learning process. A naive way to approach this is to approve samples above some low advantage threshold $a_0$ and increase that threshold over time, or to approve samples with an advantage higher than $x_0\%$ of the dataset and increase that threshold as learning progresses. However, it is challenging to set those hyperparameters across different tasks. We design a binary filter that uses a pairwise statistical T-test to create a much more task-invariant hyperparameter. We estimate the advantage of a sample $k$ times and then estimate the advantage of actions recommended by the current policy $k$ times, and only approve the dataset action for cloning if its advantage is greater than that of the current policy with statistical confidence $p_t$, where $p_t$ can be annealed from $1.0$ to high confidence $.05$. Experiments show that this does smooth the experience approval curve, but the final performance only matches that of the standard binary filter in the control tasks we study. This approach is inspired by \cite{lagoudakis2003reinforcement}, and implemented using the statistics functions in \cite{jones2001scipy}.

\paragraph{Annealed Binary Filters with Advantage Classifier Networks.} A second candidate solution to the same problem is to train an ensemble of networks that attempt to classify the advantage of actions as positive or negative. We can then use our ensemble's mean confidence and uncertainty to make informed decisions about experience approval - inspired by techniques in self-supervised learning and pseudo-labeling \cite{rizve2021defense, guo2017calibration}. This helps us deal with the noise surrounding the large fraction of samples near zero advantage because pseudo-labels have to be positive for several gradient updates before the ensemble will agree on their approval. We implement this using a similar network architecture to existing critic networks but with a sigmoid output. Once again, performance is essentially identical to the AFBC+PER baseline; the DMC tasks studied are simple enough that techniques either perform very well or fail. We think it is likely that this technique could be successfully applied to a different domain.

\paragraph{Accurate and Sample-Efficient Critics with REDQ and Weighted Bellman Updates.} The improvement of AFBC over standard BC rests on our ability to accurately estimate the advantage of $(s, a)$ pairs, which makes the method vulnerable to bias in the critic learning process. Critic overestimation and error propagation is a thoroughly investigated problem in recent work \cite{fujimoto2018addressing, lan2020maxmin, kuznetsov2020controlling, anschel2017averageddqn, kumar2020discor, agarwal2020optimistic}. REDQ \cite{chen2021randomized} trains an ensemble of critic networks on target values generated by a random subset of that ensemble and provides an effective bias-variance trade-off. It is also shown to allow for many more gradient updates per sample, something that could be useful in the offline RL context where the size of our dataset is fixed. Our main experiments are focused on high-noise datasets containing millions of samples, which makes overfitting a secondary concern, but this could be a key feature when working with small datasets. Another approach to managing overestimation error is to minimize the impact of uncertain target values in the temporal difference update. We can use REDQ's critic ensemble to estimate target uncertainty and down-weight bellman backups when $s'$ is out-of-distribution. We implement a modified version of the SUNRISE \cite{lee2020sunrise} critic loss, where we weight samples in the critic update according to a normalized uncertainty metric:
\begin{align}
    \mathcal{L}_{critic} = \mathcal{L}_{critic} * \text{softmax}(\tau \text{Std}(Q_{\{\theta_i | i \in \{0, 1, \dots n\})}(s', a' \sim \pi(s')))
\end{align}

\begin{figure}[h]
    \centering
    \includegraphics[scale=.5]{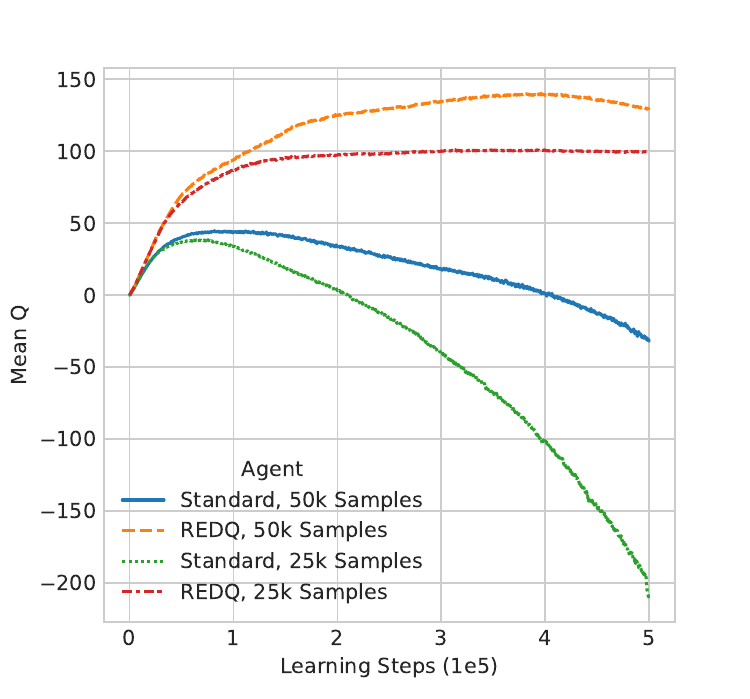}
    \caption{\textbf{Critic Ensembles in Low-Sample Datasets.} We compare the Q-function learned by a REDQ ensemble with weighted updates to that of standard clipped double-q learning in a version of \textit{``Walker, Walk" VeryBad-Expert} with just 25k and 50k transitions. The smaller dataset can cause runaway under-estimation, but the more advanced critic techniques minimize this effect.}
    \label{redq}
\end{figure}

Where $\tau$ is a temperature hyperparameter. Figure \ref{redq} displays the mean Q value of the critic networks in two sample-restricted ``Walker, Walk" datasets. Interestingly, critic bias typically results in uncontrollable \textit{under}-estimation in the AFBC context because actors are restricted to in-distribution actions, and the bias is caused by the $min$ function in the target computation rather than over-exploitation of positive bias by a policy gradient update. REDQ prevents Q-function collapse and allows for more stable learning despite a very small offline dataset.

\section{Additional Results}

This sections contains additional figures referenced in the main text that were deferred to the appendix due to the page limit.

\textbf{\begin{figure}[h!]
    \centering
    \makebox[\textwidth][c]{\includegraphics[width=1.2\linewidth]{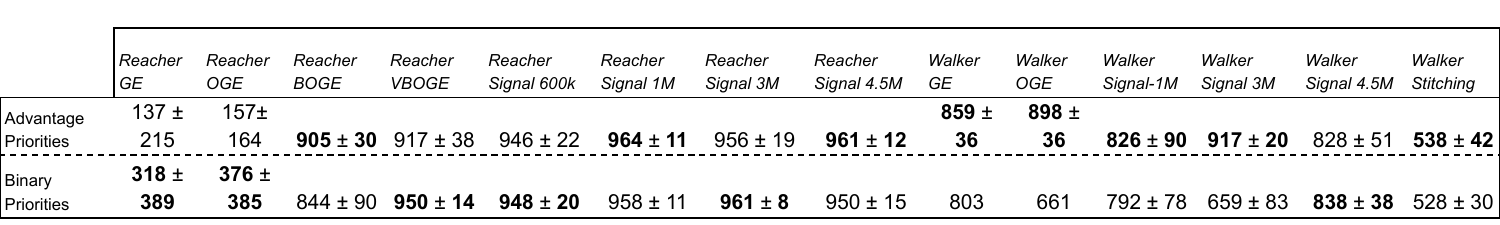}}
    \caption{\textbf{A comparison of binary and clipped advantage prioritization schemes.} The returns are very comparable. We suspect this is because the advantage values across the dataset are often small and tightly distributed; as a result, the binary priorities are actually quite similar to the true advantage weighting. Reported scores are the mean and $95\%$ confidence interval across $5$ random seeds. ``VBOGE" stands for ``Very Bad, Bad, Okay, Good, Expert" - referring to the distribution of dataset samples. In the main text this dataset is abbreviated ``\textit{VeryBad-Expert}". The same pattern applies to ``BOGE", ``OGE", and ``GE". }
    \label{fig:weight_comp}
\end{figure}
}

\label{additional-results}

\begin{figure}[h!]
    \centering
    \includegraphics[width=\linewidth]{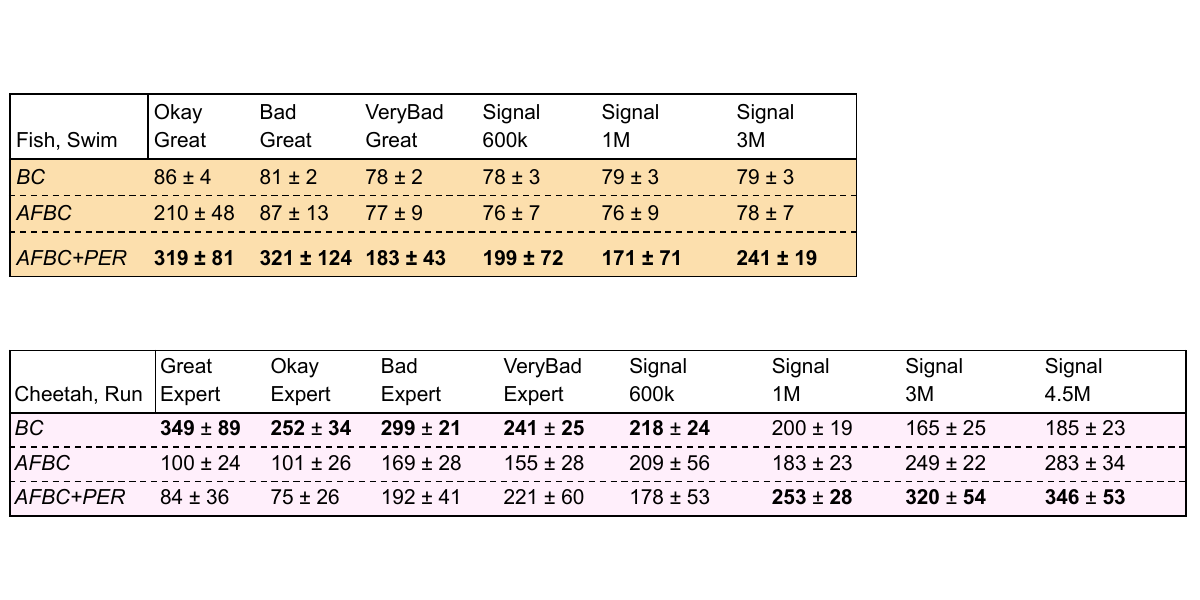}
    \caption{\textbf{AFBC in Challenging Environments where BC Fails.} There are some environments where the offline datasets appear insufficient for learning. These results use the default hparams in Table \ref{hparams_baseline}. Reported scores are the mean and $95\%$ confidence interval across $5$ random seeds.}
    \label{fig:hard_afbc}
\end{figure}

\begin{figure}[h!]
    \centering
    \includegraphics[scale=.8]{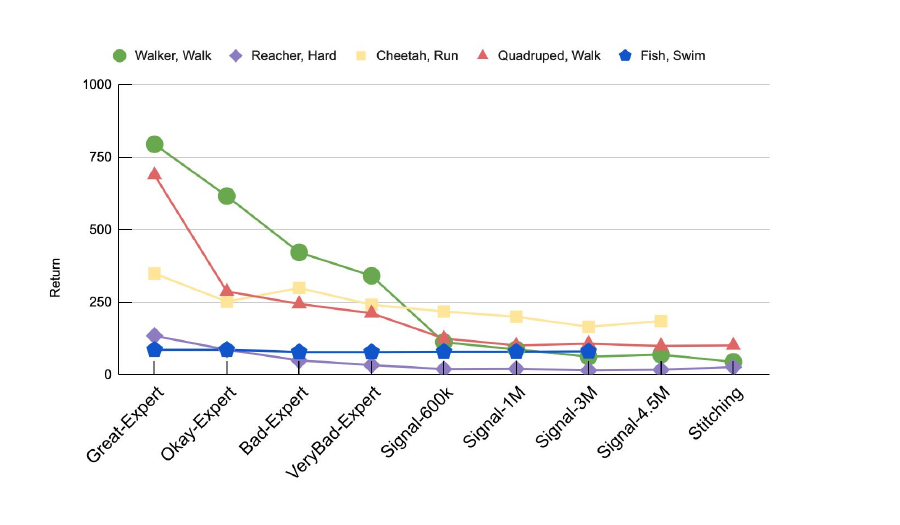}
    \caption{\textbf{Behavioral Cloning in High-Noise Datasets.} We train a BC agent on $9$ datasets across each task. The reported return is the average of $5$ random seeds. The horizontal axis is sorted by decreasing dataset quality. If BC can learn a successful policy from high-quality demonstrations, it fails to maintain its performance as the dataset fills with sub-optimal noise.}
    \label{fig:sbc_results}
\end{figure}
\end{document}